\pdfoutput=1
\documentclass[11pt]{article}

\usepackage[]{eacl2023}
\usepackage{booktabs}
\usepackage{times}
\usepackage{latexsym}
\usepackage{color}
\usepackage[T1]{fontenc}
\usepackage[utf8]{inputenc}

\usepackage{microtype}
\usepackage{amsmath}

\usepackage{graphicx}
\usepackage{stfloats}

\usepackage{multirow}

\title{FrameBERT: Conceptual Metaphor Detection with Frame \\ Embedding Learning}
\author{Yucheng Li\textsuperscript{1}, Shun Wang\textsuperscript{2}, Chenghua Lin\textsuperscript{2}\thanks{~~Corresponding author}~, Frank Guerin\textsuperscript{1}, Loïc Barrault\textsuperscript{2,3} \\
\textsuperscript{1}~Department of Computer Science, University of Surrey, UK \\
\textsuperscript{2}~Department of Computer Science, University of Sheffield, UK \\
\textsuperscript{3}~Meta AI \\
\texttt{\{yucheng.li, f.guerin\}@surrey.ac.uk}\\
\texttt{\{swang209, c.lin\}@sheffield.ac.uk}\\
\texttt{loicbarrault@meta.com}
}

\begin{document}
\maketitle
\begin{abstract}
In this paper, we propose FrameBERT, a RoBERTa-based model that can explicitly learn and incorporate FrameNet Embeddings for concept-level metaphor detection.  FrameBERT not only achieves better or comparable performance to the state-of-the-art, but also is more \textit{explainable} and \textit{interpretable} compared to existing models, attributing to its ability of accounting for external knowledge of FrameNet.

%we propose FrameBERT, a BERT-based model for conceptual metaphor detection underpinned by learning and incorporating embedding representation of semantic frames in FrameNet. 
\end{abstract}

\section{Introduction} \label{sec:introduction}

Metaphor is a pervasive linguistic device, which attracts attention from both fields of psycholinguistics and computational linguistics due to the key role it plays in the cognitive and communicative functions of language \citep{wilks1978making,lakoff1980metaphors, lakoff1993contemporary}. Linguistically, metaphor is defined as a figurative expression that uses one or several words to represent another concept given the context, rather than taking the literal meaning of the expression \citep{fass1991met}. 
For instance, in the sentence ``\textit{This project is such a \underline{headache}!}'', the contextual meaning of \textit{headache} is ``a thing or person that causes worry or trouble; a problem'', different from its literal meaning, ``a continuous pain in the head''.

Metaphor Detection presents a significant challenge as it necessitates comprehending the intricate associations between the abstract concepts embodied by the metaphoric expression and the related context. Recent studies in this field have demonstrated its potential to positively impact various Natural Language Processing (NLP) applications, including sentiment analysis \cite{cambria17sentiment,Li2023TheSO}, metaphor generation \cite{Tang2022RecentAI,Li2022CMGenAN,Li2022NominalMG}, and mental health care \cite{abd2017analysing,gutierrez-etal-2017-using-mental-health-care}. Different strategies have been proposed for modeling relevant context, including employing limited linguistic context such as subject-verb and verb-direct object word pairs \cite{gutierrez-etal-2016-literal}, incorporating a wider context encompassing a fixed window surrounding the target word \cite{do-dinh-gurevych-2016-token,mao2018word}, and considering the complete sentential context \cite{gao2018neural,choi2021melbert}.
Some recent efforts attempt to improve context modelling by explicitly leveraging the syntactic structure (e.g., dependency tree) of a sentence in order to capture important context words, where the parse trees are typically encoded with graph convolutional neural networks~\cite{le2020multi,song2021verb}. 
%However, such an encoding process is tree-dependent, making the batch operation inconvenient during optimisation. In addition, it is not uncommon that a sentence contains multiple metaphoric target words, which can easily lead to a target word being influenced by unrelated context words and relations. 

Despite the progress, we also observe the inadequacy of existing models in semantic modelling, which plays a crucial role in metaphor detection. 
For instance, it has been noted that BERT's tendency to aggregate shallow semantics instead of precise meaning, as its objective, may limit the context modelling ability~\cite{xu2020symmetric}.  
External knowledge such as FrameNet has been widely used to provide extra semantic information and has been shown useful in a wide range of NLP tasks, such as question answering~\cite{shen-lapata-2007-fsp}, machine reading comprehension~\cite{guo2020frame-for-mrc}, and identifying software requirements~\cite{alhoshan2019using}. Very recently, FrameNet has also been employed to the task of metaphor generation via learning mappings between domains, with promising results achieved~\cite{stowe2021meta-gen-fn}. 
%a different but related task to ours, to learn mappings between
However, such a valuable source of knowledge, surprisingly, has not been explored in the deep learning literature for metaphor detection. We hypothesise that incorporating external knowledge of concepts is essential for improving metaphor detection and model explainability. 

In this paper, we propose FrameBERT, a BERT-based model for conceptual metaphor detection underpinned by learning and incorporating embedding representation of semantic frames in FrameNet. 
FrameBERT directly addresses the limitation of the existing works, which solely rely on the shallow semantics captured by hand-crafted psycholinguistics features or encoded by large pre-trained language models such as BERT. 
This is achieved by explicitly learning and incorporating FrameNet embeddings into the model training process. 
To our knowledge, this is the first attempt to apply FrameNet in deep learning models for metaphor detection. 
We also leverage Metaphor Identification Procedure \citep[MIP]{group2007mip, steen2010method} and Selectional Preference Violation (SPV) \citep{wilks1975,wilks1978making} to inform our model design.
%, yielding better or comparable performance to the state-of-the-art models. 

%Second, inspired by the work in aspect-based sentiment analysis  \cite{wang2020relational}, we develop a flat, target word oriented tree structure by reshaping and pruning the ordinary dependency trees. The resulting tree representation is able to retain cleaner and more relevant context information for the target word, as well as to facilitate both batch and parallel operations. 

% We introduce a novel metaphor identification model based on two metaphor identification theories: the Selectional Preference Violation (SPV) theory \cite{wilks1975,wilks1978}, and the Metaphor Identification Procedure (MIP) theory. The SPV theory states a metaphor is identified by noticing a semantic contrast between a target word and its context. For instance,  In contrast to existing approaches, our method utilize concepts information

%We evaluate our model against the state-of-the-art models for metaphor detection. 
%Apart from the joint classification, the individual classifications for dialogue acts and topics are also compared with other baselines ~\cite{yang2016hierarchical,ji2016latent,kumar2017dialogue} {\bf [MC: Topic part now dropped.]}.
Extensive experiments conducted on four public benchmark datasets (i.e., VUA MOH-X, TroFi) show that FrameBERT can significantly improve metaphor detection for all datasets compared to our base model without exploiting FrameNet embeddings. %and the syntactic information from the unified tree structure. 
Our model also yields better or comparable performance to state-of-the-art models in Micro F1 measure. Furthermore, we show the \textit{explainable} feature of FrameBERT, attributing to its ability of extracting semantic frames from text. The code and dataset can be found at \url{https://github.com/liyucheng09/MetaphorFrame}.

\begin{figure*}[ht]
    \centering
    \includegraphics[width=0.88\textwidth]{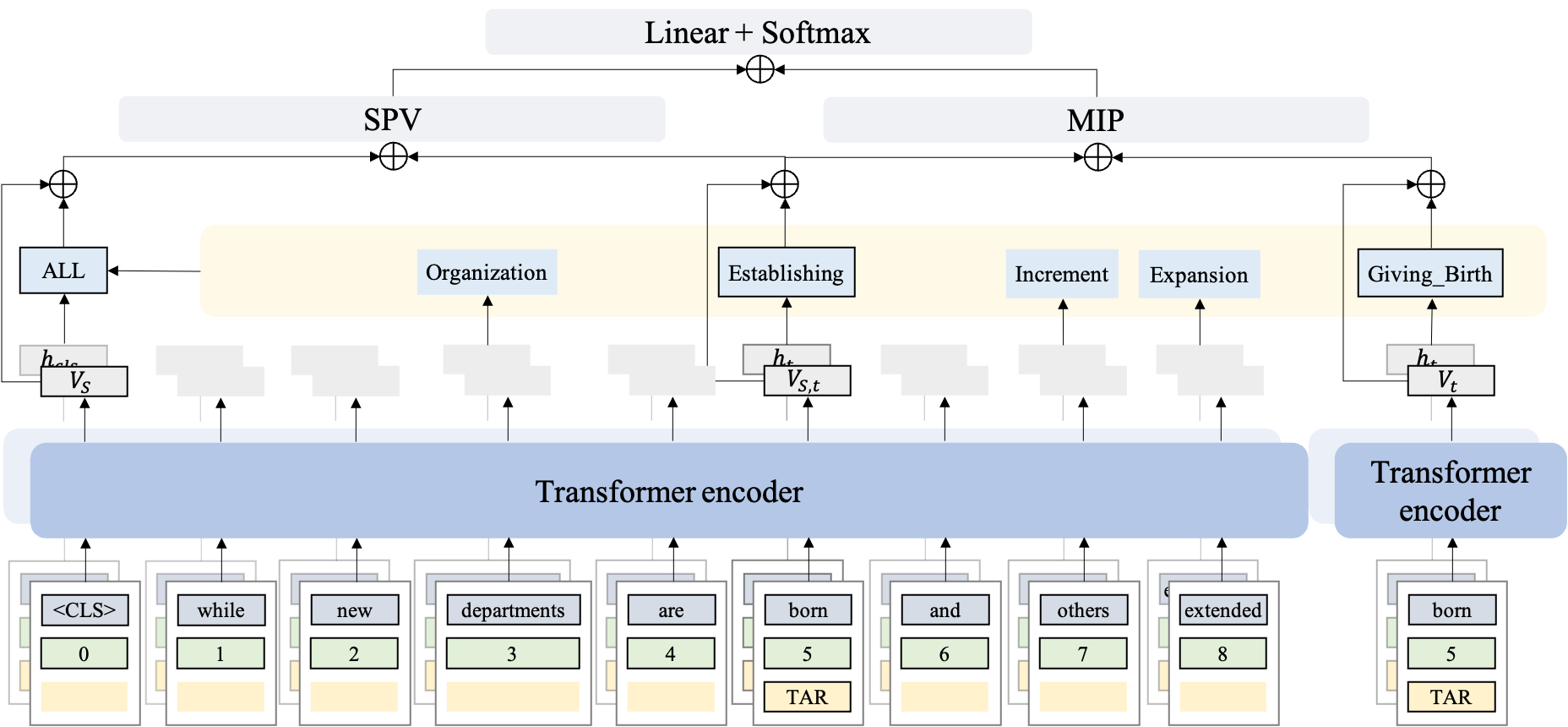}
    \caption{The overall framework. The surface encoder illustrates sentence encoder providing hidden-state representations and the insider one shows concept encoder producing concepts information. The frame embedding and hidden state embedding are concatenated to make final predictions.}
    \label{fig:framework}
\end{figure*}

\section{Model} \label{sec:methods}

We propose FrameBERT, a novel model that can explicitly learn and incorporate FrameNet embeddings for concept-level metaphor detection. 
%We also leverage  metaphor identification theories, MIP and SPV to inform our model design.  
Figure~\ref{fig:framework} illustrates the overall architecture of FrameBERT, which consists of two components: a sentence encoder (\S~\ref{semantic encoder}) and a concept encoder (\S~\ref{concepts encoder}).

%In this section, we introduce a novel framework where the MIP and SPV theories are realised in both semantic (\S~\ref{semantic encoder}) and concepts dimension (\S~\ref{concepts encoder}). Formally, given a sentence $S=(w_1, ..., w_n)$, our framework aims to predict the metaphoricity (metaphorical or literal) of the target word $w_t \in S$. Specifically, as shown in figure \ref{fig:framework}, the overall framework can be roughly divided into two parts: the semantic encoder to capture sentence's meaning, the contextual, and the literal meaning of target word; the conceptual encoder to capture concepts information for both sentence and target word. %The rest of this section is organised as follows. \S~\ref{semantic encoder} presents the semantic encoder, \S~\ref{concepts encoder} introduce the conceptual encoder and how it is been used in the overall framework.

\subsection{Sentence Encoder}
\label{semantic encoder}
Similar to the prior work \cite{choi2021melbert, song2021verb}, we develop the sentence encoder to produce the sentence encoding $\mathbf{v}_{S}$, the contextualised encoding for the target word $\mathbf{v}_{S,t}$, as well as isolated encoding for the target word $\mathbf{v}_{t}$.  
%As the MIP compares contextual meaning and literal meaning of target word, and SPV messures the gap between sentence meaning and contextual meaning of target word, three semantic representation should be modeled.
%The previous works used \textsc{cls} embedding as the sentence representation \citep{choi2021melbert, song2021verb}. 
Formally, given an input sequence $S=(w_0, ..., w_n)$, RoBERTa \cite{liu2019roberta} encodes each word into a set of contextualised embedding  vectors $\mathbf{H}=(\mathbf{h}_{cls}, \mathbf{h}_0, ..., \mathbf{h}_n)$: % network to produce represents for each word.
\begin{equation}
    \mathbf{H}=\mathrm{RoBERTa}(\mathrm{emb}_{cls}, ..., \mathrm{emb}_{n})
\end{equation}
where \textsc{cls} token is a special token used to indicate the the beginning of the  input; 
%$\mathbf{H}=(\mathbf{h}_{cls}, \mathbf{h}_0, ..., \mathbf{h}_n)$ is the output hidden states;
$\mathrm{emb}_{i}$ is the input embedding for word $w_i$ represented as %$\mathrm{emb}_{i} = \mathrm{emb}_w + \mathrm{emb}_{pos} + \mathrm{emb}_{type}$.  
%which can be derived as follows:
\begin{equation}
    \mathrm{emb}_{i} = \mathrm{emb}_w + \mathrm{emb}_{pos} + \mathrm{emb}_{type} 
\end{equation}
Here $\mathrm{emb}_{w}$ represents the word embedding, $\mathrm{emb}_{pos}$ is the position encoding for $w_{i}$, $\mathrm{emb}_{type}$  token type encoding indicating whether a word is a target or non-target word. We employ the  \textsc{cls} hidden state $\mathbf{h}_{cls}$  as the sentence representation, i.e., $\mathbf{v}_{S}=\mathbf{h}_{cls}$,   
%\begin{equation}
%\label{equa:sentence embedding}
%   \mathbf{v}_{S} = \mathbf{h}_{cls}
%\end{equation}
% is a hidden stats set of semantic neighbours of target word, where the neighbours is obtained from RoBERTa hidden states $H=(h_{0},...,h_{k})$ guided by the target-oriented dependence tree. $n$ is the length of $N_{target}$. 
the hidden states $\mathbf{h}_{t}$ of target word $w_t$  as the contextual target word embedding, i.e., $\mathbf{v}_{S,t}=\mathbf{h}_{t}$.
%\begin{equation}
%\label{equa:contextual target embedding}
%    \mathbf{v}_{S,t} = \mathbf{h}_{t}
%\end{equation}
For the isolated word embedding for $w_t$, we directly feed $w_t$ to RoBERTa in order to obtain the literal representation of the target word, i.e., $\mathbf{v}_{t} = \mathrm{RoBERTa}(\mathrm{emb}_t)$. 
% \begin{equation}
% \label{equa:literal target embedding}
%     \mathbf{v}_{t} = \mathrm{RoBERTa}(\mathrm{emb}_t)
% \end{equation}

\noindent\textbf{FrameBERT using MIP and SPV.}~~With MIP, a metaphorical word is identified by the gap between the contextual and literal meaning of a word. To incorporate MIP, we employ the contextualised $\mathbf{v}_{S,t}$ and isolated embedding $\mathbf{v}_{t}$ vectors for $w_{t}$. With SPV, a metaphorical word is identified by the semantic difference from its surrounding words, i.e., the contrast between $\mathbf{v}_{S}$ and $\mathbf{v}_{S,t}$. We formalise our incorporation of these two metaphor identification theories below. Note that $\oplus$ is an operation for vector concatenation.
\begin{align}
\label{equa:MIP embedding}
    \mathbf{h}_{MIP} & = \mathbf{v}_{t} \oplus \mathbf{v}_{S,t} \\
\label{equa:SPV embedding}
    \mathbf{h}_{SPV} & = \mathbf{v}_{S,t} \oplus \mathbf{v}_{S}
\end{align}

\subsection{Conceptual Encoder} 
\label{concepts encoder}

% \textcolor{red}{[CL: Yucheng to add some paragraphs describing your approach to concept detection.]}
One of the key contributions of our paper is that our model can explicitly learn and incorporate FrameNet Embeddings for concept-level metaphor detection. This is achieved via the conceptual encoder, where we first fine-tuning a RoBERTa model on the FrameNet \cite{fillmore2002framenet} dataset with a  objective to classify frame lables, and then join the conceptual encoder with the sentence encoder. 

%We extend the two metaphor detection theories to the conceptual dimension. For MIP we compare the literal concepts and the contextual concepts of the target word rather than  its hidden-state-meaning. For SPV we compare  the input sentence's concepts and target's concepts. %, where the original SPV theory tries to analyse the gap between the sentence meaning and the target meaning. 
% Note that in all three cases (sentence meaning, contextual meaning, literal meaning) we allow for multiple concepts to represent the meaning; we actually use a distribution.
%We use FrameNet frames to provide concept information
% : literal concepts and contextual concepts for target words and sentence concepts for input sentence in the metaphor detection process, 
%as shown in Figure \ref{fig:framework}. Before putting the conceptual information into our metaphor detection framework, we first perform a frame identification pre-training process. %, as shown in Figure \ref{fig:frame_pretrain}.

% \begin{figure*}[ht]
%     \centering
%     \includegraphics[width=0.8\textwidth]{figures/frame_pretrain.png}
%     \caption{The frame identification pre-training stage, where target word embedding are used to predict target frame and \textsc{cls} embedding predict all frames occur in the input.}
%     \label{fig:frame_pretrain}
% \end{figure*}

Given an input sentence  $S=(w_0, ..., w_n)$, we add a special token \textsc{cls} at the beginning of the sentence and apply a stack of Transformer encoder layers on the tokenised input to obtain the  contextualised hidden states for each word $\mathbf{H}=(\mathbf{h}_{cls}, \mathbf{h}_0, ..., \mathbf{h}_n)$ and the \textsc{cls} token, similar to \S~\ref{semantic encoder}.
We then leverage the contextual target word hidden states and \textsc{cls} hidden states (as sentence representation) to predict the target word's frame and all frames detected in the sentence. Formally, given \textsc{cls} hidden states $\mathbf{h}_{cls}$ and a list of contextualised target word hidden states $\mathbf{H}=(\mathbf{h}_0, ..., \mathbf{h}_k)$, we obtain the frame distribution for sentence and targets as follows:
\begin{align}
\label{equa:cls}
    & \mathbf{\hat y^f}_{cls} = \mathrm{sigmoid}(\mathbf{W}_0\mathbf{h}_{cls}+\mathbf{b}_0) \\
\label{equa:y}
    & \mathbf{\hat y^f} = \mathrm{softmax}(\mathbf{W}_1\mathbf{H}+\mathbf{b}_1)
\end{align}
where $\mathbf{W}_0$ and $\mathbf{W}_1$ are learnable parameters, $\mathbf{b}_0$ and $\mathbf{b}_1$ are bias. Note that $\mathbf{\hat y^f}_{cls}$ should be applied on all frame classes, that is compute it on each possible frame classess. We then minimise the following loss functions:
\begin{align}
 \mathcal{L}_{target} & = -\sum \mathbf{y} \log \mathbf{ \hat y^f} \\
 \mathcal{L}_{cls} & = -\sum^N\sum_{i=0}^L \mathbf{y}_i\log \mathbf{\hat y^f}_{cls} \\ & + (1-\mathbf{y}_i)\log(1- \mathbf{\hat y^f}_{cls})
\end{align}
where $N$ is the number of training samples. $L$ is number of frame labels, which means we are optimising the objective on all possible frame classes. We use $\lambda$ as a hyperparameter controling weights between two losses: $\mathcal{L}=\lambda * \mathcal{L}_{cls} + \mathcal{L}_{target}$; and we set it to 2 in our experiments.

After the pre-training stage, the conceptual encoder will provide frame information for metaphor detection. As shown in Figure \ref{fig:framework}, 
in the MIP module, we concatenate the contextualised frame embedding $\mathbf{h}_{S,t}$ and isolated frame embedding $\mathbf{h}_{t}$ of target word to $\mathbf{h}_{MIP}$ (eq. \ref{hmip}). In the SPV module, we concatenate the \textsc{cls} frame embedding $\mathbf{h_{cls}}$ and contextualised target word frame embedding $\mathbf{h}_{S,t}$ to $\mathbf{h}_{SPV}$ (eq. \ref{hspv}).
\begin{align}
\label{hmip}
    \mathbf{h}_{MIP} &= \mathbf{v}_{t} \oplus \mathbf{v}_{S,t} \oplus \mathbf{h}_{t} \oplus \mathbf{h}_{S,t} \\
\label{hspv}
    \mathbf{h}_{SPV} &= \mathbf{v}_{S} \oplus \mathbf{v}_{S,t} \oplus \mathbf{h}_{cls} \oplus \mathbf{h}_{S,t}
\end{align}
We then combine two hidden vectors $\mathbf{h}_{MIP}$ and $\mathbf{h}_{SPV}$ to compute a prediction score.
\begin{equation}
    \mathbf{\hat y} = \sigma(\mathbf{W}_T(\mathbf{h}_{MIP} \oplus \mathbf{h}_{SPV}) + \mathbf{b})
\end{equation}
Finally, we use binary cross entropy loss to train the overall framework for metaphor prediction.
\begin{equation}
        \mathcal{L} = -\sum^N_{i=0} \mathbf{y}_i\log \mathbf{\hat y}_i - (1-\mathbf{y}_i)\log(1-\mathbf{\hat y}_i)
\end{equation}

\section{Experiments}
\noindent\textbf{Dataset.}~~We conduct experiments on four public bench datasets. 
\textbf{VUA-18} \citep{leong2018report} and \textbf{VUA-20} \citep{leong2020report}, the extension of VUA-18, are the largest publicly available datasets.
The \textbf{MOH-X} dataset is constructed by sampling  sentences from WordNet \cite{miller1998wordnet}. Only a single target verb in each sentence is annotated. The average length of sentences is 8 tokens, the shortest of our three datasets. 
\textbf{TroFi} \cite{birke2006-trofi} consists of sentences from the 1987-89 Wall Street Journal Corpus Release 1 \cite{charniak2000-wsj8789}. The average length of sentences is the longest of our datasets (i.e., 28.3 tokens per sentence). At last, the concept encoder was pre-trained on \textbf{FrameNet  release 1.7} \cite{fillmore2002framenet} with about 19k, 6k, 2k annotations for training, testing and evaluation respectively.
\vspace{2mm}
\noindent\textbf{Baselines.}
\textbf{RNN\_ELMo} \citep{gao2018neural} combined ELMo and BiLSTM as a backbone model.
\textbf{RNN\_MHCA} \citep{mao2019end}: they introduced MIP and SPV into RNN\_ELMo and capture the contextual feature within window size by multi-head attention. \textbf{MUL\_GCN} \citep{le2020multi} apply a GCN based multi-task framework by modelling metaphor detection and word sense disambiguation.
\textbf{RoBERTa\_SEQ} \citep{leong2020report} is a fine-tuned RoBERTa model in sequence labeling setting for metaphor detection.
% : is used as a main pre-trained language model based baseline. This method provides a detail setting for a BERT based sequence labeling model.
%\textbf{RGAT} \citep{wang2020relational}: we adopted their GAT-based model and replaced the original sentiment task with our Metaphor task. This model take BERT as encoder before extracting features by applying attention mechanism with reshaped dependence tree. 
\textbf{MelBERT} \citep{choi2021melbert} realize MIP and SPV theories via a RoBERTa based model. 
%For \textbf{TroFi} and \textbf{MOH-X} datasets, we compare our model with sota baseline and two valuable previous works. 
\textbf{MrBERT} \citep{song2021verb} is the recent sota on verb metaphor detection based on BERT with  verb relation encoded. 

%\subsection{Overall results}
%\textcolor{red}{Since both VUA18 and VUA20 are unbalanced datasets, it is more valuable to use F1-score %as the standard for comparing the model, 
%rather than Accuracy.  
\begin{table}
\resizebox{\columnwidth}{!}{
\begin{tabular}{l|ccc|ccc}
\toprule 
 \multirow{2}{*}{Models}    &  & VUA18  &  &  & VUA20  &    \\ \cmidrule{2-7}

               & Prec & Rec  & F1 & Prec & Rec  & F1      \\ \midrule
RNN\_ELMo             & 71.6 & 73.6 & 72.6 & - & - & -          \\
RoBERTa\_SEQ          & 80.1 & 74.4 & 77.1 & 75.1 & 67.1 & 70.9         \\
MelBERT $\star$              & 79.6 & 76.4 & 77.9 & 76.4 & 68.6 & 72.3         \\
MelBERT  & 80.1 & 76.9 & 78.5  & 75.9 & 69.0 & 72.3 \\
MrBERT                  & 82.7 & 72.5 & 77.2 & - & - & -\\
FrameBERT       & 82.7    & 75.3    & \textbf{78.8*} & 79.1 & 67.7 & \textbf{73.0*} \\  \bottomrule
\end{tabular}
}
\caption{Performance comparison on VUA datasets (best results in \textbf{bold}). NB: $\star$ indicates the reproduced results of MelBERT using the original source code and setting of \cite{choi2021melbert}. * denotes our model outperforms the competing model with $p < 0.05$ for a two-tailed t-test; except MrBERT whose code is not published.}  
\label{tabel:VUA18_result}
\end{table}

\begin{center}
\begin{table}[]
\centering\small
\begin{tabular}{l|lccc}
\toprule
 & \multicolumn{1}{c}{Models} & Prec                     & Rec                      & F1                                \\ \midrule 
\multirow{4}{*}{\rotatebox{90}{TroFi}}                       & RNN\_MHCA                        & \multicolumn{1}{l}{68.6} & \multicolumn{1}{l}{76.8} & \multicolumn{1}{l}{72.4}          \\
                       & MUL\_GCN                        & \multicolumn{1}{l}{73.1} & \multicolumn{1}{l}{73.6} & \multicolumn{1}{l}{73.2}          \\
                       & MrBERT                     & \multicolumn{1}{l}{73.9} & \multicolumn{1}{l}{72.1} & \multicolumn{1}{l}{72.9}          \\\cmidrule{2-5}
                       & FrameBERT                  & \multicolumn{1}{l}{70.7} & \multicolumn{1}{l}{78.2} & \multicolumn{1}{l}{\textbf{74.2}} \\ \midrule
\multirow{4}{*}{\rotatebox{90}{MOH-X}} & RNN\_MHCA                      & 77.5                     & 83.1                     & 80.0                              \\
                       & MUL\_GCN                         & 79.7                     & 80.5                     & 79.6                              \\
                       & MrBERT                     & 84.1                     & 85.6                     & \textbf{84.2}                     \\\cmidrule{2-5}
                       & FrameBERT                 & 83.2                     & 84.4                     & 83.8                              \\ \bottomrule
\end{tabular}
\caption{Performance comparison of our method with baselines on TroFi and MOH-X (best results in \textbf{bold}). We do not perform a significance test since the code of MrBERT is not published.}
\label{tabel:trofi and moh results}
\end{table}
\end{center}
\section{Experimental Results}
\label{sec: results}
\noindent\textbf{Overall results.}~~Table~\ref{tabel:VUA18_result} shows a comparison of the performance of our model against the baseline models on VUA18 and VUA20 respectively. 
Our model outperforms all the baseline models on VUA-20, including  the state-of-the-art-model MelBERT (with $p < 0.05$ for a two-tailed t-test). For VUA-18, we outperformed all the baselines significantly including the \textit{re-produced} results for MelBERT.
% It  should be noted that we are not able to  re-produced results for MelBERT on VUA-18, despite efforts and using the original code.
Table \ref{tabel:trofi and moh results} shows the results on the MOH-X and TroFi dataset. The results shows our method beats SOTA method on TroFi benchmark and gains comparable performance on MOH-X dataset.
%where we have a slight advantage. %  It is clear from the tables that both of our enhancement methods (Reshaped\_tree and Concept) beat SOTA.}

\noindent\textbf{Ablation Study.}~~
We performed three experiments to test the effectiveness of conceptual information. First, the system is fed with shuffled conceptual embeddings in the batch during evaluation. Second, in both training and evaluation processes, we shuffle the conceptual embeddings in the batch. Third, we remove the concept fine-tuning process. In all experiments, the overall framework remains the same as the original setting.
The results are provided in Table~\ref{tabel:adversarial_results}.
Based on the results, the performance in terms of F1 drops by 13\% and 3.7\% while feeding random conceptual information in only evaluation stage and both training and evaluation stages (likely collapse into the base model) respectively, which demonstrates the extent to which the conceptual information is incorporated in the overall framework (i.e. especially when we shuffle only for evaluation). The third experiment shows the performance declines 1.2\% while removing the frame fine-tuning procedure, which proves the effectiveness of frame embedding learning.

% \noindent
% \textbf{Embedding Compound}
% \label{embedding compound}
% As discussed in \S~\ref{sec:methods}, %SPV and MIP compare the combination of [sentence, contextual\_target] and [contextual\_target, isolated\_target] respectively, only the [sentence, isolated\_target] combination is ignored. 
% %Therefore, 
% we also designed experiments to test the the [sentence, isolated\_target] combination.
% se three combinations.
% [sentence, isolated\_target] appears to be more consistent with the SPV theory because it considers the wider context of the sentence rather than focusing on the possible meanings of the target word. We refer to this as SPV* in the results.
% As we can see from Table~\ref{tabel:Compound_method}, the combination with SPV*  does not perform as well as using SPV+MIP. \textcolor{red}{[maybe delete]}
% \begin{center}
% \begin{table}[]
% \centering\small
% \begin{tabular}{l|ccc}
% \toprule
% Models    & Prec & Rec  & F1            \\ \midrule
% SPV+MIP                        & 82.7 & 74.0 & \textbf{78.6} \\
% SPV*+MIP                       & 83.7 & 72.9 & 78.0          \\ \bottomrule
% \end{tabular}
% \caption{Performance of different embedding compound types: SPV* means use sentence embedding and isolated-target embedding to replace the original SPV pair.}
% \label{tabel:Compound_method}
% \end{table}
% \end{center}
% ------------------ 5.4 -----------------
%\subsection{The Effectiveness of Concept in Assisting Concept Detection}
% \begin{center}
\begin{table}[]
\centering\small
\begin{tabular}{l|ccc}
\toprule
Models    & Prec & Rec  & F1         \\ \midrule
FrameBert                        & 82.7 & 75.3 & \textbf{78.8}  \\
rand\_in\_eval                      & 81.8  & 58.7 & 68.3          \\
rand\_in\_train\_\&\_eval                 &  79.3 & 72.6 & 75.8         \\
w/o frame fine-tuning & 79.1 & 76.3 & 77.6 \\
\bottomrule
\end{tabular}
\caption{Results of ablation study, tested on VUA18. \textit{rand\_in\_eval} represents the first experiment where the shuffle process is only performed in evaluation stage and \textit{rand\_in\_train\_\&\_eval} represents the second experiment where the shuffle process is performed in both training and evaluation stages. In \textit{w/o frame fine-tuning} experiment, we remove the frame fine-tuning process.}
\label{tabel:adversarial_results}
\end{table}
% \end{center}
\noindent\textbf{Concept Analysis.}~~ In this section, we illustrate how the proposed approach detect metaphor in an interpretable way and how well the method using frame features. We performed an exploratory analysis on 200 examples where our system had a correct classification, but MelBERT failed. The following two examples show how frame information works in the metaphor detection procedure.
The forst is a true positive example with the target word in bold: `\textit{Local people mutter and march, make speeches and throw things; staff \textbf{face} sarcasm in nearby pubs \ldots}.' Here our system had the following concepts as the literal meaning for `\textit{face}': `\textit{Body\_parts, Facial\_expression, Change\_posture}', which are more basic meanings, relating to the face as a part of the body. In contrast, contextual concepts are extracted as follows:
`\textit{Confronting\_problem, Resolve\_problem, Surviving}'. These capture well the contextual meaning of `face' in the sentence. The contextual meanings are more abstract, and the contrast between literal and contextual concepts helps the model to detect the metaphorical usage of \textit{face} here. An example of a true negative is: `\ldots
\textit{\textbf{hot} computers are slow, the warmth might damage}\ldots'. `Hot' is a word that can often be used metaphorically (e.g. hot topic, hot pants, hot properties), but in this sentence our model correctly identified it as literal and contextual concepts identified were identical: `\textit{Temperature, Fire\_Burning}'. In terms of how well our method using frame features, we measured the accuracy of the frame prediction module manually for these 200 examples, and found the correct frame label was identified in the top 3 frame label prediction for 178 of 200 examples (89\%). This indicates our method is effective extracting frame features.

%However it is not the case that all our true negatives (literals) showed the same concepts across literal and contextual meanings. Only 38\% had identical concepts, while 90\% had at least one concept in common. 
% The corresponding figures for true positives (metaphors) were 24\% and 82\%. Therefore classification cannot be done by a simple counting of common concepts between literal and metaphorical meaning. Instead, the neural network model must have learnt relationships among concepts that are indicative of metaphor and literal.

% In analysing the examples it is apparent that the contextual concepts and literal concepts identified are not always accurate; they are sometimes strange and inappropriate. Nevertheless the system seems to be able to get an advantage from their inclusion, possibly due to the difference between them in metaphorical cases. For example in one metaphorical sentence where the target word `anaemic' is used to describe a novel,
% the contextual concepts are: Medical\_conditions,	Medical\_specialties,	Toxic\_substance; the literal concepts are: Medical\_conditions,	Biological\_urge,	Medical\_specialties. Clearly the contextual concepts do not describe the contextual meaning (which should be `weak' or `poor'), but nevertheless `Toxic\_substance' does provide a contrast with the literal meanings, and it is these types of contrasts which seem to be exploited by the model.

%\subsection{Results Given Wrong Concepts}
\section{Conclusion}

We proposed FrameBERT, the first conceptual model for metaphor detection by explicitly learning and incorporating FrameNet Embeddings for concept-level metaphor detection.
Extensive experiments show that our model can yield better or comparable performance to the state-of-the-art.
% \clearpage

\section{Limitations}
This paper mainly models frame information by representation learning on the frame classification task. However, other features such as Frame Elements (FEs) and Lexical Units (LUs) in FrameNet have not been explored in this paper, where our case analysis shows these features could provide useful signals for metaphor detection. It might also be promising to explore other types of knowledge such as context graphs \cite{cheng-etal-2021-guiding} for improving metaphor detection. We leave these directions to future works. 
%This model provide a novel vision by utilising external concept knowledge to improve the performance. FrameBERT achieves SOTA performance compared with previous methods on VUA18 and VUA20. The experiments conducted on TroFi and MOH-X indicate good model transferability of FrameBERT.

% \section{Ethical Considerations}

% To copy something from existing papers. \textbf{Need to check whether this can be outside the 8-page limit.}

%\section*{Acknowledgements}


\bibliographystyle{acl_natbib}
\bibliography{reference,concepts}

% \appendix

% \section{Example Appendix}
% \label{sec:appendix}

%This is an appendix.

\end{document}